\documentclass[lettersize,journal]{IEEEtran}
\usepackage{amsmath,amsfonts}
\usepackage{algorithmic}
\usepackage{algorithm}
\usepackage{array}
\usepackage[caption=false,font=normalsize,labelfont=sf,textfont=sf]{subfig}
\usepackage{textcomp}
\usepackage{stfloats}
\usepackage{xurl}
\usepackage{verbatim}
\usepackage{subcaption}
\usepackage{graphicx}
\usepackage{caption}
\usepackage{xcolor}
\usepackage{makecell}
\usepackage{multirow}
\usepackage{tablefootnote}
\usepackage{multicol}
\usepackage{latexsym}
\usepackage{booktabs}
\usepackage{tabularx} 
\usepackage{cite}
\usepackage[
    colorlinks=true,
    citecolor=blue,
    linkcolor=black,
    urlcolor=blue
]{hyperref}
\usepackage{orcidlink}
\newcolumntype{Y}{>{\raggedright\arraybackslash}X}

\begin{document}

\title{Rationale Behind Human-Led Autonomous \\ Truck Platooning}

\author{Yukun Lu$^{1*}$\,\orcidlink{0000-0003-3259-5945}, Chenzhao Li$^{2}$, Xintong Jiang$^{1}$, Qiaoxuan Zhang$^{1}$

\textsuperscript{1}Dept. of Mechanical Engineering, University of New Brunswick, Fredericton, NB E3B 5A3, Canada

\textsuperscript{2}Faculty of Interdisciplinary Studies, University of New Brunswick, Fredericton, NB E3B 5A3, Canada
}

\maketitle

\begin{abstract}
    Autonomous trucking has progressed rapidly in recent years, transitioning from early demonstrations to OEM-integrated commercial deployments. However, fully driverless freight operations across heterogeneous climates, infrastructure conditions, and regulatory environments remain technically and socially challenging. This paper presents a systematic rationale for human-led autonomous truck platooning as a pragmatic intermediate pathway. First, we analyze 53 major truck accidents across North America (2021-2026) and show that human-related factors remain the dominant contributors to severe crashes, highlighting both the need for advanced assistance/automated driving systems and the complexity of real-world driving environments. Second, we review recent industry developments and identify persistent limitations in long-tail edge cases, winter operations, remote-region logistics, and large-scale safety validation. Based on these findings, we argue that a human-in-the-loop (HiL) platooning architecture offers layered redundancy, adaptive judgment in uncertain conditions, and a scalable validation framework. Furthermore, the dual-use capability of follower vehicles enables an evolutionary transition from coordinated platooning to independent autonomous operation. Rather than representing a compromise, human-led platooning provides a technically grounded and societally aligned bridge toward large-scale autonomous freight deployment.
\end{abstract}

\section{Introduction}

Freight transportation forms the backbone of modern supply chains, yet heavy-duty trucking continues to face persistent safety, operational, and workforce challenges. Human-related factors remain a dominant contributor to severe truck accidents, while increasing freight demand intensifies pressure on drivers and logistics systems. At the same time, rapid advances in sensing, computation, and artificial intelligence have accelerated the development of assisted and autonomous driving technologies for commercial vehicles. Although fully driverless trucking has progressed from experimental pilots to early commercial deployments, significant technical, regulatory, and societal uncertainties remain.

This paper examines a central transitional question: should humans be fully removed from the driving loop in heavy-duty freight operations, or does a hybrid paradigm offer a more resilient pathway? We argue that human-led autonomous truck platooning represents a pragmatic and strategically balanced intermediate architecture between conventional driver assistance and fully driverless operation.

The contributions of this paper are threefold:

(1) We provide an accident-driven rationale for assisted and autonomous trucking by analyzing 53 major truck crashes across North America (2021–2026). The analysis identifies recurring behavioral, environmental, and mechanical factors, demonstrating that while human error remains dominant, many crash modes are heterogeneous and context-dependent, complicating immediate full autonomy deployment.

(2) We present a structured review of the recent evolution of the autonomous trucking industry, identifying three developmental phases: early feasibility demonstrations (2015–2016), consolidation and systems maturation (2020–2023), and emerging commercial driverless deployment (2025–2026). By examining OEM integration, redundancy architectures, and corridor-based scaling strategies, we clarify both progress and persistent long-tail limitations.

(3) We propose a systems-level rationale for human-led platooning as an intermediate deployment architecture. We argue that embedding professional human oversight within platoon leadership provides layered redundancy, adaptive judgment in edge cases, and a scalable validation envelope. Furthermore, we articulate the dual-use capability of follower vehicles as a pathway toward gradual autonomy expansion without technological lock-in.

By integrating accident analysis, industry trajectory assessment, and architectural reasoning, this paper positions human-led truck platooning not as a compromise, but as a strategically structured bridge toward large-scale autonomous freight deployment.

\begin{figure}[ht]
    \centering
    \includegraphics[width=\linewidth]{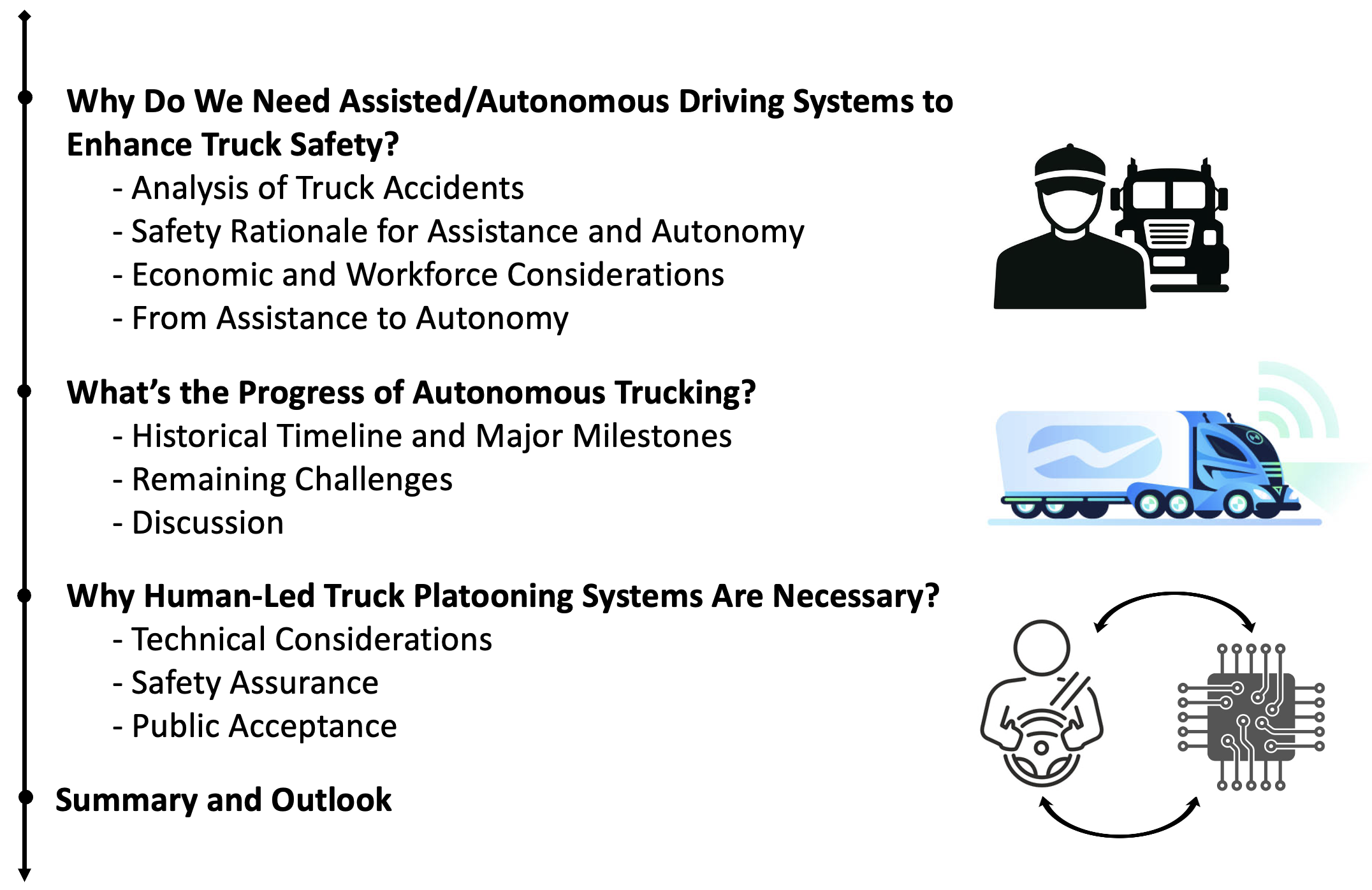}
    \caption{Overview of structure and guiding research questions.}
\end{figure}

\section{Why Do We Need Assisted/Autonomous Driving Systems to Enhance Truck Safety?}

\subsection{Analysis of Truck Accidents}
To assess the major causes of serious and fatal truck crashes, we collected and analyzed 53 major truck accidents that occurred across North America between 2021 and 2026. The data were sourced from official investigation reports (e.g., NTSB, RCMP) and credible news media, covering a wide range of scenarios including highway crashes, weather-related incidents, and complex interactions with other vehicles. A summary of the causes is presented in Figure X.
Truck driver-related factors account for the largest proportion (58.5\%), followed by factors involving other vehicles (24.5\%). Mechanical failures and adverse weather/environment each contributed to less than 10\% of the incidents. This distribution underscores that human errors, either by truck drivers or by drivers of surrounding vehicles, remain the predominant cause of major truck accidents.

\subsubsection{Driver-Related Factors}

A closer examination of the 31 accidents attributed to truck drivers reveals recurring patterns of unsafe behavior. These can be grouped into three major categories: unsafe driving habits, inexperience and poor judgment, and violations of traffic laws.

\textbf{Unsafe Driving Habits:}
A closer examination of unsafe driving habits reveals several recurring patterns. Fatigue and inattention were contributing factors in multiple crashes where drivers failed to respond to slowed or stopped traffic ahead, often after long hours behind the wheel. For example, the June 2021 crash on I-65 in Alabama occurred when a commercial truck driver did not react to a stationary queue \cite{NTSB_HWY21MH009}. Similarly, a 2022 Virginia crash was officially attributed to driver fatigue and hours-of-service violations \cite{NTSB_HIR2405}. Following too closely was also identified as a critical issue, as evidenced by a 2026 Missouri crash involving a tractor-trailer rear-ending an agricultural tractor due to insufficient following distance \cite{TheTrucker_MissouriCrash2026}. Additionally, prolonged left-lane occupancy, although not explicitly cited in the dataset remains a concern. Highway patrol reports frequently note that trucks lingering in the left lane can provoke unsafe passing maneuvers from other vehicles, increasing collision risk \cite{Carscoops_SemiHabit2025}.

\textbf{Inexperience and Poor Judgment:}
Inexperience and poor judgment frequently manifest as improper handling in adverse weather conditions, where speeding or abrupt maneuvers on wet, icy, or snow-covered roads lead to loss of control. The February 2025 I-80 tunnel fire in Wyoming was triggered by a pickup losing control on wet/icy roads, followed by a semi jackknifing inside the tunnel \cite{NTSB_HWY25MH004}. In Michigan, January 2025, a tractor-trailer lost control on the icy road, crossed the center line, and collided with an oncoming vehicle \cite{MichiganThumb_SebewaingCrash2025}. A Texas crash in January 2026 was attributed to hydroplaning \cite{People_I35Crash2026}. Misjudgment of vehicle dynamics also contributes significantly to accidents, particularly when overloading or failure to reduce speed on curves causes rollovers. In British Columbia, November 2024, a semi-trailer lost control on a downhill curve and overturned; the driver was cited for driving without due care \cite{BCRCMP_SemiTrailerCollision2024}. Furthermore, operating at speeds inappropriate for prevailing conditions remains a persistent problem. A 2025 South Carolina crash occurred when a truck overturned after hitting a barrier, with investigators explicitly noting "speed too fast for conditions" as a contributing factor \cite{TheSun_ChickenSpill2025}.

\textbf{Violations of Traffic Laws:}
Violations of traffic laws represent a significant category of driver-related factors contributing to major truck accidents. Speeding was a factor in multiple crashes, including the 2021 Chiapas incident involving a heavily overloaded truck \cite{NYTimes_MexicoTruckCrash2021} and a 2025 multi-vehicle chain collision in Texas where the semi failed to control speed \cite{MRT_OdessaI20Crash2025}. Impaired driving also led to severe consequences, as demonstrated by a December 2024 crash in Port Coquitlam, BC, where a tractor-trailer driver was charged with impaired operation and exceeding the blood alcohol limit \cite{BCRCMP_PortCoquitlam2024}. Failure to obey traffic controls frequently resulted in intersection and work zone accidents, with drivers ignoring stop signs or failing to yield. Examples include the October 2023 Wisconsin crash at a stop sign \cite{WSAW_WoodCountyCrash2023} and two separate incidents in Alberta (November and December 2025) where semi-trucks ran stop signs \cite{RCMP_BrooksCollision2025}. Wrong-way driving and improper overtaking further contributed to fatal collisions, such as a Florida Turnpike crash in August 2025 involving an illegal U-turn that caused a head-on collision \cite{CBS12_TurnpikeCrash2025}. A June 2025 crash in British Columbia further illustrates the dangers of illegal overtaking in the opposing lane on curved mountain roads, where a semi truck rolled over after attempting to pass at excessive speed \cite{cdllife_2025_brake_failure_topple}. Work zone violations remain a persistent issue as well, with a 2024 North Carolina crash involving a semi that failed to slow in a work zone, rear-ending a queue of vehicles \cite{NTSB_HWY24MH010}.


\subsubsection{Factors Involving Other Vehicles}

In 13 accidents, the actions of other road users were the primary cause. Common scenarios included passenger vehicles crossing the center line (e.g., January 2025 in New Brunswick \cite{rcmp_nb_2026_collision}), SUVs failing to yield (August 2024 in Kentucky \cite{wave3_2024_i64_crash}), and wrong-way driving (June 2025 in New York \cite{timesunion_2025_i87_wrong_way}). These cases highlight that truck drivers often have limited time to react to unpredictable behaviors of smaller vehicles.

\subsubsection{Mechanical Factors}

Although less frequent, mechanical failures like brake malfunctions (Mexico, December 2021 \cite{cbc_2021_mexico_toll}) or tire blowouts (Iowa, January 2026 \cite{who13_2026_i80_jasper}) also contributed to severe accidents. Improper pre-trip inspections were directly responsible for the November 2024 incident near Chilliwack, BC, where a wheel detached from a dump truck and struck an oncoming SUV, an event deemed entirely preventable with a proper pre-trip inspection \cite{rcmp_bcarchives_2024_dump_wheel}. Additionally, failure to secure cargo, while not explicitly documented in the dataset, remains a known contributor to rollovers. The 2025 Michigan milk tanker rollover may have involved such factors \cite{michigansthumb_2025_milk_tanker_rollover}.

Overall, this study underscores that driver errors are not monolithic but stem from a range of behaviors, some habitual, some judgment-based, and some illegal. Addressing these diverse root causes requires intelligent systems that can not only detect imminent hazards but also provide guidance to correct unsafe patterns before they lead to accidents.

\subsection{Safety Rationale for Assistance and Autonomy}
Advanced driver assistance systems (ADAS) and higher levels of automation aim to mitigate these human limitations through continuous sensing, algorithmic decision-making, and real-time actuation. Technologies such as automatic emergency braking, lane-keeping assistance, adaptive cruise control, blind-spot monitoring, and electronic stability control already demonstrate measurable safety benefits. 


Autonomous driving systems extend these capabilities by integrating multi-modal sensor fusion to achieve 360-degree environmental perception. Unlike human drivers, such systems maintain continuous vigilance without fatigue or distraction and operate with deterministic reaction times. In addition, they are typically designed with redundant braking, steering, and power subsystems to enhance fault tolerance and operational safety.


\subsection{Economic and Workforce Considerations}

Beyond safety, the economic rationale for autonomous trucking is substantial. Its primary value proposition lies in higher asset utilization and labor reconfiguration. By operating beyond human Hours-of-Service (HOS) constraints, autonomous trucks can function nearly continuously, limited mainly by refueling, charging, and maintenance. For long-haul corridors, this increased operational time translates into greater revenue-generating mileage per vehicle per day, faster asset turnover, improved delivery cycle times, and enhanced return on capital investment.
Although autonomy introduces additional costs, including hardware integration, high-performance computing, redundancy requirements, and operational infrastructure, these investments may be offset by productivity gains and reduced dependence on long-haul driver availability.

At the workforce level, the trucking industry faces persistent driver shortages and high turnover rates, particularly in long-haul segments that require extended time away from home. Autonomous systems can help reconfigure labour roles via shifting drivers toward first-mile/last-mile operations, supervision of multiple vehicles, or remote fleet oversight, rather than eliminating human participation entirely.

\subsection{From Assistance to Autonomy}
It is important to view assisted and autonomous systems as part of a continuum. Driver assistance technologies already reduce crash severity and frequency. Higher levels of automation build upon these systems by progressively transferring dynamic driving tasks from human operators to validated software stacks operating within defined operational design domains.

The need for assisted and autonomous driving systems in trucking therefore arises from a combination of safety imperatives, economic pressures, workforce constraints, and supply-chain efficiency demands. As freight volumes increase and logistics networks grow more complex, relying solely on human endurance and judgment may become increasingly unsustainable. Carefully engineered autonomy offers a pathway toward safer, more productive, and more resilient freight transportation systems.

\section{What’s the Progress of Autonomous Trucking?}

\begin{figure}[ht]
    \centering
    \includegraphics[width=\linewidth]{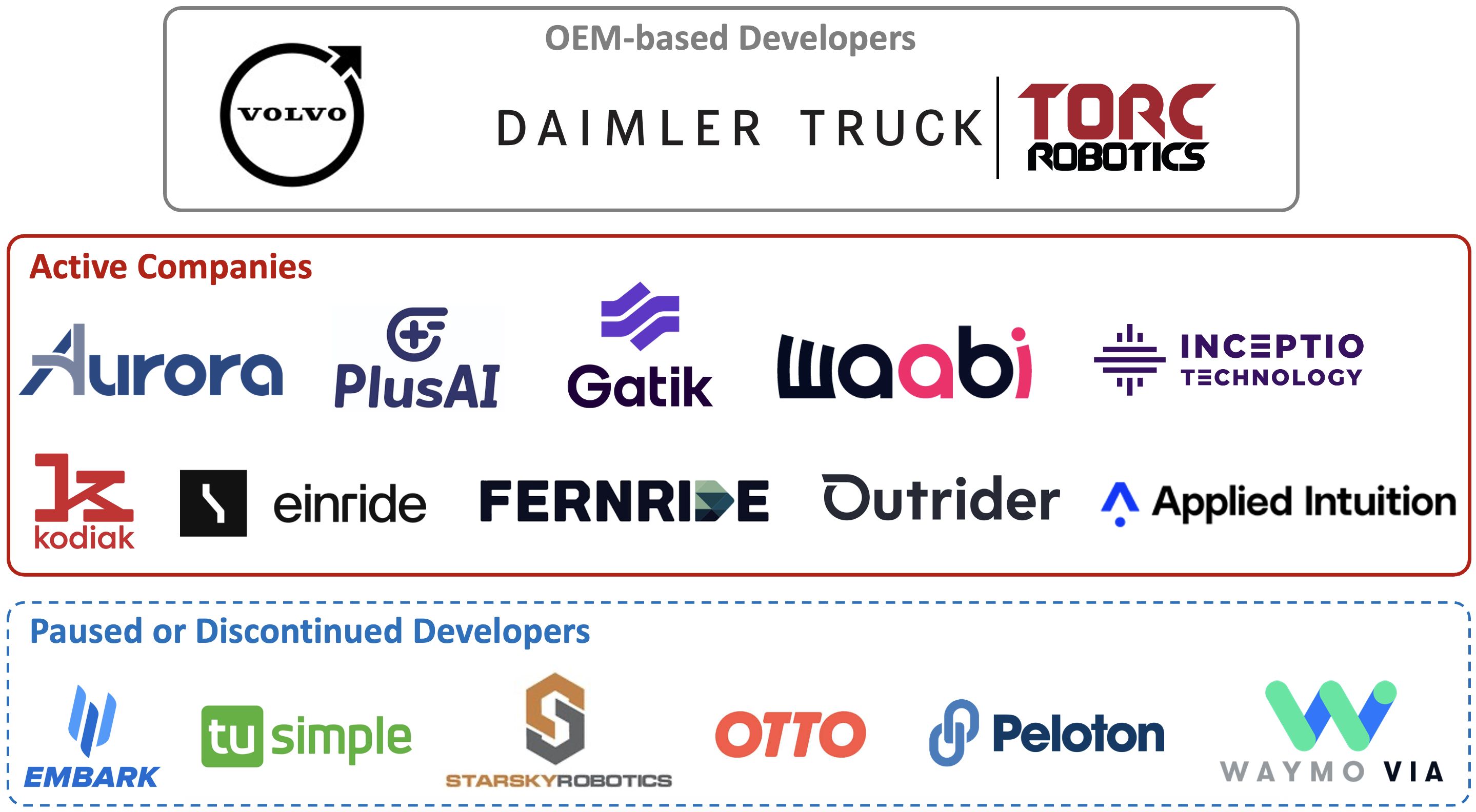}
    \caption{Major developers in the autonomous trucking sector.}
    \label{fig:companies}
\end{figure}

The commercial autonomous trucking sector has experienced a turbulent yet transformative half-decade. The industry has undergone consolidation, technological maturation, and early commercial deployment, signaling that large-scale autonomy may be closer than previously anticipated. While several high-profile startups have collapsed, others have transitioned from pilot testing to revenue-generating operations, as listed in Figure~\ref{fig:companies}. 

\subsection{Historical Timeline and Major Milestones}

The modern trajectory of autonomous trucking can be traced back to 2015, when Nevada granted a license to the Freightliner Inspiration Truck for autonomous highway testing \cite{globocam2015inspiration}. This marked one of the first officially recognized prototype milestones for heavy-duty highway automation. In 2016, Otto and Anheuser-Busch completed a widely publicized autonomous beer delivery in Colorado \cite{cbc2016budweiser}. While technologically significant, these early demonstrations primarily established feasibility under controlled highway conditions rather than scalable commercial readiness.
This period between 2015 and 2016 demonstrated that long-haul highway automation was technically achievable. However, these efforts remained largely experimental and promotional in nature. Durable commercialization frameworks that could cover safety validation, fleet integration, and regulatory alignment were not yet mature.

In 2021, Germany enacted a legal framework permitting Level 4 \cite{sae2021j3016_202104} autonomous vehicles within defined operating domains, representing one of the first comprehensive national regulatory structures for high-level automation \cite{taylorwessing2026legal}. Meanwhile, OEM partnerships deepened, most notably between Aurora and Volvo \cite{Volvo2021Aurora}, reflecting a transition toward factory-integrated autonomy rather than aftermarket retrofits.

The early wave of speculative capital that fueled numerous autonomous trucking startups inevitably led to consolidation. Companies such as Otto (2018) \cite{sawers2018uber_otto_shutdown}, Starsky Robotics (2020) \cite{templeton2020starsky}, Peloton Technology (2021), Embark (2023) \cite{AppliedIntuition2023Embark}, and TuSimple (2024) \cite{tusimple2024} ceased operations. Waymo Via significantly reduced or paused its autonomous trucking initiatives to prioritize robotaxi development. Uber shut down its autonomous trucking division in 2018 to focus on other areas of autonomy. These setbacks raised concerns about the long-term viability of the sector. However, rather than signaling collapse, the failures appear to represent a correction phase, filtering out unsustainable business models and leaving technologically and strategically stronger players. In 2020, Aurora acquired Uber ATG \cite{uber2020atg}, consolidating talent and intellectual property within fewer, better-capitalized players. This marked a shift from demonstration-driven momentum to long-term systems engineering.



The year 2025 marked a decisive turning point.
Aurora launched what is described as the first fully autonomous commercial heavy-duty trucking service operating on public roads in Texas \cite{aurora2025commercial}. Aurora’s driverless trucks began completing regular freight deliveries between Dallas and Houston without human operators onboard. Within months, Aurora expanded operations to Phoenix and launched a second route between Fort Worth and El Paso after surpassing 100,000 driverless miles on public roads \cite{aurora2025fortworth_elpaso}. Notably, Aurora also secured a commercial agreement with Detmar Logistics to deploy 30 autonomous trucks in 2026 for hauling frac sand in the Permian Basin, marking the first public-road autonomous transport of proppants \cite{aurora_detmar_2025}. These trucks are expected to operate over 20 hours per day, illustrating the economic value proposition of 24/7 driverless freight.
Canadian company Waabi represents a new generation of AI-first autonomy developers. In October 2025, Waabi integrated its Waabi Driver into the Volvo VNL Autonomous truck platform, built on NVIDIA DRIVE AGX Thor and NVIDIA DRIVE AGX Hyperion 10 architectures \cite{waabi_volvo_future_autonomous_trucking_2025}. Waabi emphasizes a vertically integrated autonomy architecture built around end-to-end interpretable AI models designed for transparency and generalization. Its system incorporates full hardware and software redundancy across critical components, including power distribution, steering, braking, and onboard computing, to ensure safe operation under fault conditions. In addition, Waabi deploys comprehensive onboard fault monitoring and management systems that continuously assess vehicle and autonomy health, enabling appropriate fallback strategies when anomalies are detected. These capabilities are complemented by a cloud-based autonomy mission management platform that supports fleet-level supervision, deployment orchestration, and continuous performance monitoring. Waabi explicitly rejects reliance on remote real-time vehicle control due to latency risks. Instead, remote systems provide high-level supervisory input while safety-critical decision-making remains onboard. This architecture reflects an important maturation of safety philosophy in autonomous trucking.
In parallel, Kodiak’s commercialization emphasizes \emph{customer-owned} fleets and a licensing model. For Atlas Energy Solutions, Kodiak highlights driverless RoboTruck operations hauling proppant in the Permian Basin \cite{kodiak2025_two_trucks}.
International and PlusAI announced a factory-built Level 4 program based on NVIDIA DRIVE AGX Hyperion platform \cite{international_plusai_2025}, signalling deeper OEM integration. 

Entering 2026, the focus has shifted toward scaling deployment. Aurora validated a 1,000-mile autonomous lane exceeding human hours-of-service limits and announced plans to deploy more than 200 driverless trucks. 
Kodiak, in collaboration with Bosch, advanced the manufacturing of production-grade redundant autonomous platforms \cite{kodiak_bosch_2026}. These developments indicate a clear transition from pilot validation to fleet-scale operational readiness.

Another significant development is the shift from retrofit solutions to factory-integrated autonomy. 
Volvo Autonomous Solutions (VAS), a business unit of the Volvo Group, has made significant progress in autonomous trucking over the past several years. 
Between 2022 and 2023, Volvo Autonomous Solutions formally introduced its Autonomous Transport Solution concept, targeting hub-to-hub freight operations in North America. The focus has been on structured highway corridors where operational domains can be clearly defined and optimized for Level 4 autonomy. VAS adopted a partnership-driven model, collaborating with logistics providers and freight operators to integrate autonomous trucks into real supply chain networks. Strategic collaborations with companies such as DHL Supply Chain and Uber Freight have allowed Volvo to evaluate operational, economic, and safety performance in commercial freight environments. 
Recent years have seen the transition from controlled pilots to structured commercial freight operations in the United States. Volvo VNL Autonomous trucks equipped with partner-developed Level 4 driving systems have operated on long-haul corridors such as Dallas-Houston, Fort Worth-El Paso, and other Sunbelt highway routes.
Meanwhile, Volvo has begun integrating sensors, computing hardware, and redundant control systems directly on production lines. For example, Volvo has announced the production of autonomous-ready truck platforms, including the Volvo FH Autonomous in Europe and the Volvo VNL Autonomous in North America. A key component of this ecosystem strategy has been close technical collaboration with autonomy software developers, including Aurora Innovation and Waabi. Rather than developing a fully proprietary autonomous stack, Volvo integrates validated autonomy systems into purpose-built Volvo truck platforms.
International partnered with PlusAI to commercialize SAE Level 4 trucks using PlusAI’s SuperDrive virtual driver and NVIDIA’s DRIVE AGX Thor platform. PlusAI reports over six million real-world autonomous driving miles, leveraging end-to-end AI models capable of adapting across diverse routes and geographies \cite{}.
Daimler/Torc are targeting a 2027 commercial launch and have selected Innoviz for short-range LiDAR for series production of autonomous Freightliner Cascadia trucks. This indicates that autonomy components are moving into production validation phases rather than remaining experimental add-ons.


Overall, the evolution of autonomous trucking can be characterized in three phases: initial feasibility demonstrations (2015–2016), consolidation and systems engineering maturation (2020–2023), and the emergence of regular driverless commercial operations with OEM-integrated production programs (2025–2026). The industry’s progress has shifted from high-visibility stunts to rigorous safety cases, redundancy architectures, supply-chain integration, and scalable fleet deployment.


\subsection{Remaining Challenges}
Despite technological progress, widespread commercial deployment is not guaranteed. Regulatory uncertainty at both state and federal levels remains a significant barrier. While testing continues to refine reliability and operational robustness, legislation may ultimately determine the pace of large-scale adoption.
Nevertheless, the convergence of AI acceleration, OEM partnerships, production-grade hardware, and early commercial deployments suggests that hundreds of autonomous trucking routes across North America could become operational by the end of the decade \cite{jones2025market}.

\subsubsection{Technical Limitations}
The hardest remaining technical problems are not ``keeping the lane'' but sustaining safe behavior in the long tail: complex construction zones, degraded markings, unusual merges, and adverse weather plus sensor contamination. 

Autonomous trucks rely on multi-modal sensor suites, typically combining LiDAR, radar, and camera systems. While sensor fusion improves environmental awareness, each sensing modality has inherent vulnerabilities. Cameras are sensitive to glare, low-light conditions, and adverse weather such as fog or heavy rain. LiDAR performance may degrade in snow, dust, or heavy precipitation, and long-range detection accuracy can be affected by atmospheric scattering. Radar, while robust to weather, provides lower spatial resolution and may struggle with object classification.
Heavy-duty trucking further amplifies these challenges due to extended sensing requirements. Long stopping distances necessitate reliable object detection at ranges exceeding several hundred meters. Highway-speed perception must be both long-range and highly reliable.

Autonomous systems perform well in structured and statistically common scenarios but remain challenged by rare, unpredictable, or ambiguous situations, often referred to as long-tail events. Examples include unusual vehicle behaviors, erratic human drivers, unexpected construction zones, temporary traffic control patterns, or debris falling from other vehicles.
Machine learning-based perception systems are trained on large datasets, yet no dataset can comprehensively represent all possible real-world configurations. Generalization to unseen environments remains imperfect. Even highly advanced end-to-end AI models may struggle with rare combinations of environmental factors, lighting conditions, and dynamic interactions. This creates residual uncertainty in safety-critical decision-making, particularly when operating beyond narrowly defined operational design domains (ODDs).

Most autonomous trucking deployments today are limited to well-defined highway corridors with relatively predictable traffic patterns. Expanding beyond these constrained operational domains introduces exponential complexity. Urban environments, complex interchanges, rural roads without clear lane markings, and cross-border regulatory differences increase system uncertainty.
Mapping and localization systems also require high-definition maps for precise positioning. Maintaining and updating such maps across large geographic regions is resource-intensive. Road construction, lane reconfiguration, and temporary signage can rapidly invalidate pre-mapped assumptions.

\emph{Winter-time operation} represents one of the most significant technical difficulties for autonomous trucking systems, particularly in northern regions such as Canada and the northern United States. Snow, ice, freezing rain, blowing snow, and extreme low temperatures introduce compounded challenges across perception, localization, vehicle dynamics, and hardware reliability.
Winter conditions further complicate vehicle dynamics and motion planning. Reduced tire-road friction on snow or ice increases stopping distances and alters steering response characteristics. Heavy-duty trucks, due to their mass and articulated trailers, are particularly susceptible to jackknifing and trailer sway under low-traction conditions. Autonomous control algorithms must adapt braking, acceleration, and steering commands to account for variable friction coefficients, often in real time. Estimating road friction accurately remains a non-trivial problem, especially when black ice is present and visually indistinguishable from dry pavement.
From a perception standpoint, snowfall and snow accumulation degrade sensor performance in multiple ways. Camera systems may experience reduced visibility due to low contrast between snow-covered roads and lane markings. Lidar signals can scatter or attenuate in heavy snowfall, producing noisy or spurious returns. Ice and slush accumulation on sensor housings may partially occlude fields of view, requiring active cleaning mechanisms. Radar remains relatively robust to precipitation, but it lacks the spatial resolution required for fine-grained object classification and lane-level perception. In severe winter storms, overall sensor fusion confidence may drop, increasing uncertainty in object detection and tracking.
Localization is also significantly affected. Many autonomous systems rely on high-definition maps combined with lane markings and roadside features for precise positioning. When roads are covered by snow, lane boundaries and pavement edges may become invisible, invalidating key map-based references. Snowbanks can alter road geometry, and plowed snow may temporarily narrow lanes or shift traffic patterns. In such scenarios, localization algorithms must rely more heavily on inertial measurement units (IMUs), GNSS, and radar-based feature detection, all of which have their own limitations in precision and drift.
Extreme cold temperatures introduce hardware reliability concerns. Low temperatures can affect battery performance, sensor electronics, wiring insulation, and actuator responsiveness. Lidar and camera systems may require heating elements to prevent frost formation. Hydraulic and pneumatic brake systems may experience slower response times. Ensuring fail-operational redundancy under prolonged exposure to sub-zero temperatures increases engineering complexity and energy consumption.
Operationally, winter weather also introduces unpredictable human behavior. Surrounding drivers may brake suddenly, lose control, or deviate from lanes. Snowplows, emergency vehicles, and temporary road closures alter normal traffic patterns. Construction zones may be obscured by snow, and temporary signage can be partially covered. These dynamic and often ambiguous situations fall into the category of long-tail events that remain challenging for AI-based decision-making systems.
For freight operators in northern climates, reliable year-round operation is essential for economic viability. Therefore, winter capability is not merely an edge case but a critical requirement for widespread deployment. Addressing winter-time challenges demands advances in robust sensor design, adaptive control algorithms, friction estimation techniques, sensor self-cleaning mechanisms, thermal management systems, and winter-specific safety validation frameworks.
In summary, winter driving presents a multi-layered systems challenge for autonomous trucks, affecting perception, localization, motion control, hardware reliability, and human interaction simultaneously. Achieving dependable performance under severe winter conditions remains a key barrier to fully scalable autonomous freight operations.

Given the impossibility of perfect pre-deployment validation, humans serve as a crucial fallback for unanticipated situations. When autonomous systems encounter scenarios outside their operational design domain, they must either hand control to a human or execute a minimum-risk maneuver (such as pulling over and stopping). While minimum-risk maneuvers provide a basic safety layer, they can create their own hazards, for example, a stopped truck on a highway shoulder presents risks to other motorists. 
Human drivers, by contrast, can navigate novel situations using general intelligence and contextual understanding. They can interpret ambiguous cues, make judgment calls about acceptable risk, and execute complex maneuvers that automated systems cannot. This fallback capability is so valuable that many safety experts consider it essential for the foreseeable future.

\subsubsection{Public Trust}
Beyond formal regulatory requirements, autonomous trucking must earn public trust to achieve widespread acceptance. High-profile incidents involving autonomous vehicles have damaged public confidence and created political headwinds. Maintaining human presence in the loop may be essential for sustaining social license during the transition period.
The trust-building function is subtle but crucial. Gradual introduction of automation with visible human oversight allows the public to become comfortable with the technology and provides opportunities to demonstrate safety before removing humans entirely.

\subsection{Discussion}
Widespread autonomy across heterogeneous climates and nationwide networks is feasible but hinges on (a) provable safety at scale, (b) standardization of regulatory acceptance (especially for unattended roadside events), and (c) an economically sustainable equilibrium where autonomy’s added capital and operations costs are outweighed by utilization and safety gains. 

In the near term, the most plausible trajectory is corridor-by-corridor scaling in regions with favorable weather, infrastructure, and policy. Public guidance from leading players points to (i) increasing driverless miles and routes, (ii) next-gen hardware cost-downs, and (iii) fleet counts moving from dozens to hundreds, provided safety performance holds and regulatory accommodations expand.

\section{Why Human-Led Truck Platooning Systems Are Necessary?}

Despite these technological breakthroughs, a fundamental question persists: \emph{should humans be completely removed from the driving loop?} 

This section argues that for the foreseeable future, a human-in-the-loop (HiL) paradigm for autonomous trucking is not merely preferable but necessary. The reasons span technical considerations, safety assurance, and societal acceptance. Although autonomous technologies have achieved impressive milestones, the transition period, where professional drivers, semi-autonomous systems, legacy infrastructure, and evolving regulations coexist, requires careful integration of human oversight rather than wholesale replacement. In this intermediate period, hybrid intelligence can offer a more resilient pathway than full autonomy alone.

\begin{figure*}
    \centering
    \includegraphics[width=0.75\linewidth]{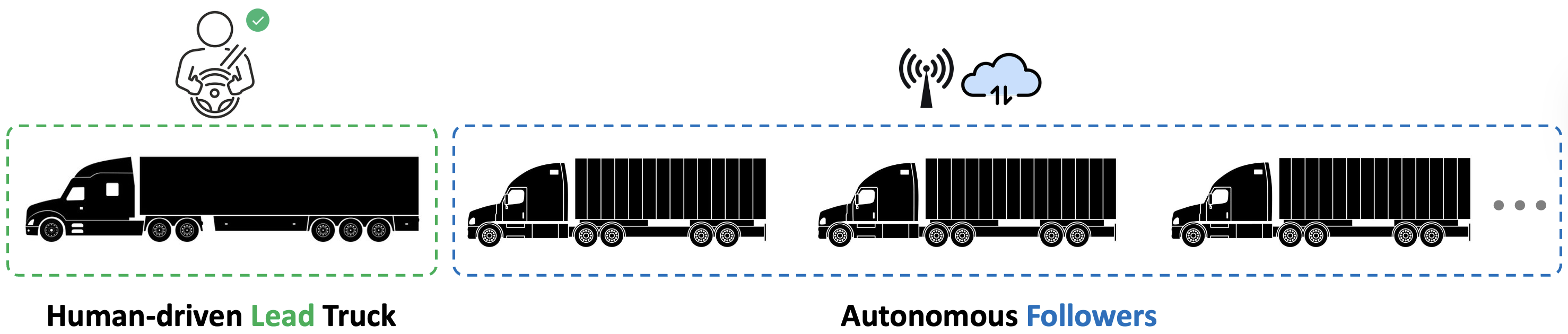}
    \caption{A sketch of the human-led truck platooning system.}
    \label{fig:placeholder}
\end{figure*}

\subsection{Technical Considerations}

Heavy-duty trucking operates in environments that frequently expose systems to edge cases: ambiguous road geometry, temporary construction zones, informal traffic negotiations, unexpected obstacles, and context-dependent right-of-way decisions. Professional truck drivers possess tacit knowledge accumulated over thousands of operational hours, particularly in interpreting subtle environmental cues and anticipating risks.

Human-led platooning systems allow automation to leverage this expertise rather than discard it. By embedding the lead vehicle under human supervision, follower vehicles can benefit from high-level human judgment while automation handles longitudinal control, spacing, and coordination. This hybrid approach enables the system to remain adaptable in rare or poorly modeled edge scenarios.

\subsubsection{Operations in Remote Regions}

Heavy-duty freight operations frequently take place in geographically remote regions, including northern corridors, resource extraction routes, and sparsely populated highways. These environments present distinct challenges: limited digital infrastructure, reduced roadside support, inconsistent connectivity, extreme weather exposure, and extended emergency response times.

Fully driverless systems operating in such contexts must independently manage rare but high-consequence events without immediate human support. In contrast, human-led platooning preserves local decision-making capability within the operational environment itself. The presence of a professional driver in the lead vehicle provides contextual interpretation, adaptive judgment, and immediate fallback capability when unexpected hazards arise.

Moreover, remote regions often exhibit degraded mapping coverage and dynamic road conditions, such as temporary gravel segments, wildlife crossings, or informal traffic behaviors. Human-in-the-loop supervision improves robustness under these uncertain and evolving conditions. As a result, human-led platooning offers a more resilient deployment strategy for remote freight corridors where environmental variability and operational risk remain high.

\subsubsection{Wintertime Operations}

Wintertime driving represents a particularly challenging domain for autonomous systems. Snow cover can obscure lane markings, degrade LiDAR returns, reduce camera contrast, and introduce highly nonlinear tire–road interactions. For heavy-duty trucks, the combination of high mass and reduced friction dramatically increases stopping distance and lateral instability risk.

A human-led platooning approach offers adaptive risk assessment during such degraded sensing conditions. Human drivers can interpret subtle contextual cues such as snow texture, wind drift patterns, and visibility changes that remain difficult to encode in purely algorithmic systems. By maintaining human oversight in adverse weather, platooning systems can preserve operational efficiency while mitigating risks associated with sensor uncertainty and reduced traction.

\subsubsection{Dual-Use Capability and Scalable Autonomy}

Human-led platooning architectures enable follower vehicles to operate under two distinct but complementary modes: coordinated platoon-following and independent autonomous driving. In platoon mode, follower trucks rely on structured leader guidance and V2V coordination to simplify perception and decision-making. However, the same perception, planning, and control stack can be incrementally trained and validated to support fully independent operation.

This dual-use capability provides a scalable development pathway. Platooning reduces environmental complexity during early deployment phases, allowing follower systems to learn structured highway behaviors under reduced uncertainty. As system maturity increases, these followers can transition toward standalone self-driving trucks capable of operating outside the platoon context.

Such an architecture avoids technological lock-in. Instead of designing follower vehicles that are permanently dependent on a lead truck, the system preserves long-term autonomy potential while maintaining near-term safety through human-led oversight. This evolutionary pathway balances immediate deployability with future full-autonomy objectives.

\subsection{Safety Assurance}

Safety validation remains one of the central challenges in autonomous trucking. Demonstrating statistical safety superiority over human drivers requires extensive real-world exposure across diverse operational conditions. 
Human-led platooning provides a pragmatic safety envelope during this validation phase. Keeping a trained driver in the loop, particularly in the lead vehicle, can:
\begin{itemize}
    \item Mitigate risk during edge-case encounters,
    \item Provide redundancy against perception or planning failures,
    \item Enable controlled fallback strategies, and
    \item Facilitate continuous system learning under supervised conditions.
\end{itemize}

This layered redundancy aligns with established safety engineering principles in aviation and industrial automation, where human supervision complements automated subsystems. Rather than replacing drivers outright, platooning distributes cognitive and control responsibilities across human and machine agents to enhance overall system resilience.

\subsection{Public Acceptance}

Beyond technical feasibility, public trust plays a decisive role in the deployment of autonomous freight systems. Large trucks already command heightened safety sensitivity. Fully removing drivers may amplify perceived risk among other road users and regulators.

Human-led platooning provides a socially and psychologically gradual transition toward higher levels of automation. The visible presence of a professional driver in the lead vehicle reassures other road users while allowing automation to demonstrate reliability over time. This incremental adoption pathway may reduce resistance from labor stakeholders and ease regulatory approval processes.


\section{Summary and Outlook}

This paper examined the safety, technological, and societal rationale for human-led autonomous truck platooning. While autonomous trucking has progressed from experimental pilots to early commercial deployment, significant challenges remain in long-tail edge cases, winter operation, remote-region logistics, and large-scale safety validation. 

In this transitional phase, human-led platooning offers a pragmatic and resilient pathway: it combines automated coordination with professional human judgment, provides layered redundancy for safety assurance, and supports gradual public and regulatory acceptance. Rather than delaying autonomy, a human-in-the-loop paradigm enables scalable deployment while preserving long-term potential for full driverless freight operations.

Looking forward, the most realistic near-term trajectory for autonomous freight appears to be corridor-by-corridor scaling under clearly defined ODDs. Human-led platooning can serve as an enabling intermediate stage, supporting continuous system validation while maintaining operational safety in complex environments. As autonomy technologies mature and safety cases strengthen, the degree of human involvement may progressively decrease. Until that point, hybrid intelligence that combines algorithmic precision with human contextual reasoning offers one of the most resilient pathways toward large-scale autonomous freight transportation.

\bibliographystyle{IEEEtran}
\bibliography{ref}

\end{document}